\documentclass[runningheads]{llncs}
\usepackage[dvipsnames]{xcolor}
\usepackage{pgfplots}      
\usetikzlibrary{patterns}  
\usetikzlibrary{pgfplots.groupplots} 
\usepackage[T1]{fontenc}
\usepackage{multirow}
\usepackage{comment}
\usepackage{amsmath}
\usepackage{tikz}
\usepackage{pgfplots}
\usepackage{pgfplotstable}
\pgfplotsset{compat=1.7}
\usepackage{pgf-pie}
\usetikzlibrary{patterns}
\usepackage{wheelchart}
\usepackage{wrapfig}
\usepackage{subcaption}
\usepackage{hyperref}
\usepackage{makecell}

\usepackage{graphicx}
\begin{document}
\title{Large Language Models for Education and Research: An Empirical and User Survey-based Analysis}
%
%
\author{Md Mostafizer Rahman\inst{1}\orcidID{0000-0001-9368-7638} \and
Ariful Islam Shiplu\inst{2}\orcidID{0009-0009-4198-9023} \and
Md Faizul Ibne Amin \inst{3}\orcidID{0009-0001-0722-3536} \and Yutaka Watanobe \inst{3}\orcidID{0000-0002-0030-3859} \and Lu Peng \inst{1}\orcidID{0000-0003-3545-286X}}
\authorrunning{M. M. Rahman et al.}
%
\institute{Tulane University, New Orleans, LA, USA \and
Dhaka University of Engineering \& Technology, Gazipur, Bangladesh \and
The University of Aizu, Aizu-Wakamatsu, Fukushima, Japan\\
\email{mrahman9@tulane.edu, shipluarifulislam@gmail.com, fiamin029@gmail.com, yutaka@u-aizu.ac.jp, lpeng3@tulane.edu}}
\maketitle              
\begin{abstract}

Pretrained Large Language Models (LLMs) have achieved remarkable success across diverse domains, with education and research emerging as particularly impactful areas. Among current state-of-the-art LLMs, ChatGPT and DeepSeek exhibit strong capabilities in mathematics, science, medicine, literature, and programming. In this study,  we present a comprehensive evaluation of these two LLMs through background technology analysis, empirical experiments, and a real-world user survey. The evaluation explores trade-offs among model accuracy, computational efficiency, and user experience in educational and research affairs. We benchmarked these LLMs performance in text generation, programming, and specialized problem-solving. Experimental results show that ChatGPT excels in general language understanding and text generation, while DeepSeek demonstrates superior performance in programming tasks due to its efficiency- focused design. Moreover, both models deliver medically accurate diagnostic outputs and effectively solve complex mathematical problems. Complementing these quantitative findings, a survey of students, educators, and researchers highlights the practical benefits and limitations of these models, offering deeper insights into their role in advancing education and research.

\end{abstract}
\section{Introduction}

Natural Language Processing (NLP) has rapidly evolved into one of the most influential areas of artificial intelligence (AI), driving progress in applications such as speech recognition, named entity recognition, machine translation, question answering, text summarization, and program code generation. Early advances were shaped by recurrent neural networks (RNNs), whose sequential structure enabled them to capture dependencies across tokens. Variants such as long short-term memory (LSTM) and gated recurrent units (GRU) \cite{multilingual10354513} alleviated the vanishing gradient problem and improved context modeling. Nevertheless, RNN-based architectures struggled to scale to extremely long-range dependencies \cite{rahman2020neural}, leading to inefficiencies in training and difficulty in modeling global semantic relationships. The emergence of the Transformer \cite{transformer11020722} architecture marked a paradigm shift in NLP by overcoming the fundamental limitations of RNNs. Transformers utilize a self-attention mechanism that enables efficient modeling of dependencies across all tokens in a sequence, regardless of their positional distance \cite{vaswani2017attention}. This breakthrough became the foundation of modern large language models (LLMs), enabling unprecedented fluency and contextual reasoning in text generation.

The development of ChatGPT by OpenAI, launched in 2022, marks a significant milestone in the evolution of LLMs. Initially based on the GPT-3 architecture and later enhanced with GPT-4, ChatGPT employs a combination of supervised fine-tuning, reward modeling, and reinforcement learning from human feedback (RLHF) \cite{rahman2023chatgpt}. This training approach was developed to address the unpredictable and undesirable outputs common in earlier models, including responses that lacked coherence, contained harmful or biased language, or presented inaccurate information \cite{kim2025chatgpt}. By integrating human feedback into its learning process, ChatGPT effectively improved the safety, reliability, and contextual coherence of its responses. Since its release, ChatGPT has become a widely adopted conversational AI capable of generating fluent, human-like dialogue across an extensive range of topics and applications.

In contrast, DeepSeek emerges in the NLP domain with a specific emphasis on knowledge retrieval, semantic search, and structured question answering. Initially introduced as an effort to design efficient yet powerful Transformer-based models and employs a mixture-of-experts (MoE) architecture that dynamically activates specialized sub-models for different tasks, reducing redundant computations and improving scalability \cite{puspitasari2025deepseek}. This design not only lowers the hardware requirements compared to conventional large-scale models but also enhances efficiency in training and inference. DeepSeek focuses on precise and context-aware information retrieval and is well-suited for domain-specific applications that require accuracy and efficient access to relevant data \cite{rahman2025chatgpt}. Furthermore, DeepSeek leverages advanced techniques such as reinforcement learning to automate and refine human feedback and achieves competitive performance on complex reasoning and mathematical tasks while keeping computational costs significantly lower than those of larger generalist models.

Although both ChatGPT and DeepSeek are based on Transformer architectures, they are optimized for different objectives. ChatGPT excels in conversational text generation, enabling applications such as creative writing, code generation, tutoring, and interactive learning, while DeepSeek prioritizes high-precision knowledge access, making it valuable for research assistance, information retrieval, and technical question answering. Together, they complement each other: ChatGPT provides conversational adaptability, and DeepSeek ensures factual accuracy, offering a combined toolkit for enhancing educational and research environments. Both models are increasingly applied in education and research. ChatGPT supports students and educators by generating lecture materials, presentations, and problem sets, while DeepSeek improves efficiency in literature reviews, structured knowledge extraction, and specialized query responses. 
These technologies highlight both opportunities, including accessibility, personalization, and efficiency, and risks, such as misinformation, academic dishonesty, and over-reliance on AI.



To the best of our knowledge, no comprehensive study has systematically compared ChatGPT and DeepSeek across the dual dimensions of technical performance and user perception in education and research contexts. Existing evaluations largely focus on single-model case studies or narrowly defined application domains. This paper addresses that gap by presenting a comprehensive comparative analysis of ChatGPT and DeepSeek. We benchmark their effectiveness in programming education, research, and academic tasks; complement these results with a user survey of students, educators, and researchers; and provide a balanced discussion of opportunities, limitations, and strategies for the responsible integration of LLMs into educational and research ecosystems.



\section{Related Work}
Recent research highlights the transformative impact of AI-generated content, including text, code, images, and mathematical solutions, across disciplines. 
Models like ChatGPT and DeepSeek have emerged as key tools, enhancing task efficiency, enabling innovative workflows, and challenging traditional methodologies, while demonstrating measurable quality through both quantitative metrics and human evaluation.

LLMs have revolutionized NLP by setting new performance standards across diverse benchmarks and tasks while demonstrating unprecedented ability to learn from massive datasets and generate coherent, contextually relevant text \cite{chowdhery2023palm}, \cite{mulia2023usability}. Their applications span critical domains such as biomedical research, where models like ChatGPT have shown potential in tasks including named entity recognition and question answering, but they have limitations in accuracy compared to specialized models such as Med-PaLM \cite{chen2023extensive}, \cite{singhal2022large}. Similarly, DeepSeek-R1 offers promising capabilities for healthcare applications through efficient and interpretable reasoning but faces challenges related to input limitations and ethical considerations \cite{zhou2025large}, \cite{zhang2025method}. In educational contexts, concerns about the impact of LLMs on academic integrity and learning practices have been raised, with research advocating for detection methods and stronger safeguards against misuse \cite{cotton2024chatting}, \cite{shijaku2023chatgpt}. Furthermore, efforts to enhance LLM training through data augmentation techniques such as AugGPT have shown improvements in model performance with limited data \cite{dai2025auggpt}, while evaluations of ChatGPT in text summarization reveal competitive results relative to fine-tuned models and highlight both its capabilities and differences from human-generated content \cite{yang2023exploring}.


Recent studies have demonstrated notable success of LLMs in code structure building, repair, classification, generation, and summarization. For instance, Siam et al. \cite{siam2024programming} evaluated top AI coding assistants such as ChatGPT, Gemini, AlphaCode, and Copilot and highlighted their progress across multiple programming languages while stressing the need for improved reliability and ethical use. Liu et al. \cite{liu2024refining} analyzed over 4,000 ChatGPT-generated programs and revealed both the potential and limitations of AI-driven code generation, including notable improvements through self-repair capabilities. Zhu et al. \cite{zhu2024deepseek} introduced DeepSeek-Coder-V2, an open-source MoE code language model that matches or surpasses GPT4-Turbo in coding, reasoning, and math tasks, supporting 338 programming languages with extended context length. Complementarily, Lu et al. \cite{lu2024deepseek} presented DeepSeek-VL, a vision-language model capable of processing high-resolution images for real-world tasks while maintaining strong language understanding. Additionally, Jalil et al. \cite{jalil2023chatgpt} found ChatGPT generated correct or partially correct answers for about 44\% of software testing questions and demonstrated its practical utility albeit with some limitations.

Furthermore, LLMs have been applied to mathematical methods, engineering calculations and questionnaire generation; however, challenges remain in generating novel mathematical equations and solving complex problems \cite{frieder2023mathematical}, \cite{khan2024chatgpt}, \cite{pham2024chatgpt}. Recent advancements highlight how generative AI tools powered by LLMs are transforming education, particularly in STEM fields. Riley et al. \cite{riley2024solving} demonstrated ChatGPT’s effectiveness in supporting personalized STEM learning by solving simultaneous equations, while Koceski \cite{koceski2025exploring} showed its capability in accurately solving ordinary differential equations, with performance depending on problem complexity and prompt clarity. Oh et al. \cite{oh2023effective} found that structured, step-by-step prompting enabled ChatGPT to correctly solve quadratic problems 91\% of the time and enhanced its potential as a personalized learning tool. Additionally, Chew \cite{chew2023overcoming} introduced the Peter Chew Method and enabled AI to overcome traditional limitations by directly using roots of quadratic equations beyond conventional approaches. In comparative studies, Jahin et al. \cite{jahin2025evaluating} reported that DeepSeek-R1 outperformed other LLMs in mathematical reasoning benchmarks, with Gemini 2.0 Flash noted for speed, while \cite{doscomparative} revealed DeepSeek-R1’s superior algebraic manipulation over ChatGPT on complex theoretical physics problems, although both struggled without visual inputs and required external tools for full computation.

\section{Background Technology Analysis}

This section provides an overview of the technological foundations of ChatGPT and DeepSeek, focusing on their evolution, architectural design, computational costs, operational mechanisms, and performance characteristics. Table~\ref{comparison} summarizes key distinctions between the two models, highlighting their core architectures, primary objectives, training data emphasis, domain-specific advantages in coding and mathematics, as well as their accessibility through deployment platforms and API support.

\begin{table}[t]
\caption{Comparison of ChatGPT and DeepSeek across key technological dimensions.}
\centering
\begin{tabular}{l|c|c}
\hline \hline
\textbf{Aspect} & \textbf{ChatGPT} & \textbf{DeepSeek} \\ \hline \hline
Core Architecture & GPT-based Transformer & Customized Transformer (MoE) \\ \hline
Primary Objective & Conversational Generation & Structured Reasoning \\ \hline
Training Data Focus & Broad Web-scale Text & Code- and Math-intensive Corpora \\ \hline
Key Strength & Fluency and Accuracy & Logical Reasoning and Efficiency \\ \hline
Domain Advantage & General-purpose + Coding Support & Strong in Coding and Mathematics \\ \hline
Deployment & Cloud-based, API Access & Limited; On-premises Options \\ \hline \hline
\end{tabular}
\label{comparison}
\end{table}


\subsection{ChatGPT}
ChatGPT, developed by OpenAI, represents one of the most widely adopted implementations of the Generative Pre-trained Transformer (GPT) family of LLMs. 
The model’s evolution has been marked by progressively larger and more capable variants: GPT-2 (2019) demonstrated fluent text generation at scale, while GPT-3 (2020) introduced 175 billion parameters and exhibited emergent few-shot and zero-shot learning capabilities. ChatGPT, introduced in late 2022, builds upon GPT-3.5 and later GPT-4, integrating advances in alignment and controllability. A critical development is its training via RLHF \cite{ouyang2022training}, where supervised fine-tuning on curated datasets is augmented with reward models trained from human preference annotations. It mitigates the issues of incoherence, factual inaccuracy, and unsafe outputs that earlier LLMs frequently exhibited. Technically, ChatGPT processes an input sequence $X=\{x_1, x_2, x_3, \cdots, x_n\}$
by embedding tokens into high-dimensional vectors, which are then transformed through stacked layers of self-attention and feed-forward networks. The model predicts the conditional probability of the next token as: $P(x_t | x_{<t};\theta)=\text{softmax}(Wh_t)$, where $ h_t$ is the contextual representation derived from multi-head attention and
$W$ is the output projection matrix. 
From an operational standpoint, ChatGPT’s training leverages large-scale distributed computing with billions of parameters optimized across massive internet-scale corpora. 
Beyond conversational dialogue, ChatGPT has demonstrated utility in educational content generation, programming assistance, scientific summarization, and creative writing

\subsection{DeepSeek}
DeepSeek represents a new class of efficient reasoning-focused large language models designed to balance performance with computational scalability. Its architecture integrates a MoE framework with Reinforcement Learning (RL), enabling competitive accuracy on mathematics, science, and coding benchmarks while substantially lowering training and inference costs compared to dense LLMs \cite{dai2024deepseekmoe,liu2024deepseek}.  DeepSeek uniquely employs a transparent chain-of-thought (CoT) process, performing step-by-step logical reasoning before answering. The model evolved from DeepSeek-V1 (67B parameters, Jan 2024) to V2 (236B parameters, Jun 2024), which introduced multi-headed laden attention and MoE for efficiency, and then to V3 (671B parameters, Dec 2024), integrating RL and advanced GPU load balancing. To further enhance scalability, DeepSeek employs knowledge distillation techniques, producing compact variants such as DeepSeek-R1-Zero, which transfer reasoning capabilities from larger models to smaller, resource-efficient versions. This strategy facilitates broad deployment in environments with limited hardware resources, without significant loss in accuracy or reasoning depth.

\section{Experimental and User Survey Results}
In this section, we present a comparative evaluation of ChatGPT, DeepSeek, and human-generated content across multiple domains. The experiments combine quantitative benchmarks with expert validation, followed by a user survey designed to capture perceptions of model effectiveness, challenges, and quality trade-offs.

\subsection{Academic and Research Affairs}
Generative AI (GAI) has demonstrated strong capabilities across a wide range of STEM and applied domains. To systematically evaluate ChatGPT and DeepSeek, we designed task-specific experiments spanning mathematics, NLP, scientific reasoning, medical applications, and programming. \textbf{Mathematics.} Both models were tasked with solving a classical Fourier series expansion problem. ChatGPT and DeepSeek independently generated step-by-step solutions, which were subsequently reviewed by subject-matter experts. Both models produced mathematically valid and complete derivations, demonstrating reliability in calculus-based tasks. 
\textbf{Language Tasks.} In NLP, both ChatGPT and DeepSeek performed consistently well in grammar correction, paraphrasing, and abstractive summarization. Their outputs were judged against human-written references and evaluated using metrics such as ROUGE and BLEU. 
\textbf{Scientific Problem Solving.} In physics, chemistry, and biology, both LLMs were tested on structured problem sets. A chemistry expert validated that both models achieved 100\% correctness on the tested problems, demonstrating their utility in domain-specific reasoning when clear prompts are provided. \textbf{Medical Applications.} Both models were further evaluated on diagnostic and reporting tasks. On average, they achieved approximately 90\% accuracy when benchmarked against gold-standard clinical references. 
\textbf{Programming.} We evaluated ChatGPT and DeepSeek on a curated set of Codeforces problems spanning difficulty ratings from 800 to 2200 (Table~\ref{program}). ChatGPT solved most problems successfully but occasionally failed due to wrong answers (WA), time limit exceeded (TLE), or compile errors (CE). DeepSeek, by contrast, achieved a perfect success rate across all evaluated problems. To quantify output quality, AI-generated solutions were compared with \textit{human solutions} using ROUGE-L, BELU, and BERTScore metrics. DeepSeek’s code solutions aligned most closely with human baselines, achieving the highest ROUGE-L (0.55), BLEU (0.51), and BERTScore (0.84) on the 800-rated task, with strong results maintained on higher-rated problems (900 and 1300). These findings collectively underscore the versatility of LLMs. ChatGPT demonstrates strengths in fluency, creativity, and user-centric explanation, while DeepSeek excels in structured reasoning, computational efficiency, and programming reliability. They highlight the potential of complementary integration for education, research, and technical domains.

\begin{table}[h]
    \caption{Performance comparison of ChatGPT and DeepSeek on Codeforces programming problems of varying difficulty levels.}
    \renewcommand{\arraystretch}{1.3} 
    \setlength{\tabcolsep}{8pt} 
    \begin{tabular}{c||p{.9cm} p{.6cm}  p{.6cm} p{.8cm}||p{.8cm} p{.2cm} p{.6cm} p{.8cm}}
        \hline\hline
        \multirow{2}{*}{\textbf{Problem}} & \multicolumn{4}{c||}{\textbf{ChatGPT}} & \multicolumn{4}{c}{\textbf{DeepSeek}} \\
        \cline{2-9}
        &  \textbf{Success} & \textbf{Fail}  & \textbf{BLEU} & \textbf{BERT} & \textbf{Success} & \textbf{Fail}  & \textbf{BLEU} & \textbf{BERT} \\
        \hline\hline
        \href{https://codeforces.com/problemset/problem/2072/A}{Problem-A}  & 5 & 0 & 0.34 & 0.77 & 5 & 0  & 0.51 & 0.84 \\
       
        \hline
        \href{https://codeforces.com/problemset/problem/2072/B}{Problem B}  & 5 & 0 & 0.18 & 0.71 & 5 & 0  & 0.33 & 0.79 \\
        \hline
        \href{https://codeforces.com/problemset/problem/2072/C}{Problem C}  & 4 & 1(WA)  & 0.12 & 0.71 & 5 & 0 & 0.16 & 0.78 \\
        \hline
        \href{https://codeforces.com/problemset/problem/2072/D}{Problem D}  & 4 & 1(CE) & 0.27 & 0.77 & 5 & 0  & 0.25 & 0.81 \\
        \hline
        \href{https://codeforces.com/problemset/problem/2072/E}{Problem E}  & 3 & 2(WA, TLE)  & 0.10 & 0.72 & 5 & 0  & 0.23 & 0.73 \\
        \hline
        \href{https://codeforces.com/problemset/problem/2072/F}{Problem F}  & 5 & 0 & 0.23 & 0.77 & 5 & 0  & 0.33 & 0.81 \\
        \hline
        \href{https://codeforces.com/problemset/problem/2072/G}{Problem G}  & 5 & 0 &  0.14 & 0.72 & 5 & 0  & 0.18 & 0.73 \\
        \hline
    \end{tabular}
    \vspace{-3mm}
    \caption*{\textbf{Note:} The difficulty levels of Problems A–G are 800, 900, 1200, 1300, 1500, 1700, and 2200, respectively.  WA = Wrong Answer, TLE = Time Limit Exceeded, CE = Compile Error.}

    \label{program}
\end{table}

\subsection{User Survey}

\begin{wrapfigure}{r}{0.5\textwidth}   
\vspace{-10mm}
  \includegraphics[width=1\linewidth]{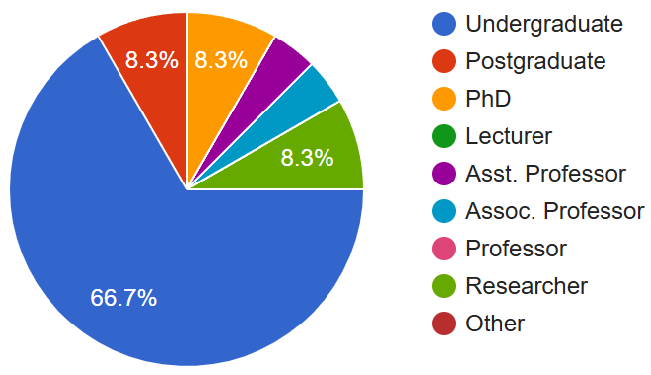}
  \caption{Survey participants by academic background}
  \label{participants_bacground}
\end{wrapfigure}

To complement the experimental evaluation, we conducted a user survey designed to assess how GAI technologies support core domains such as education, research, programming, and practical applications. The survey sought to capture perceived strengths, limitations, and expectations from a diverse set of stakeholders, including students, faculty members, researchers, and professionals. Key themes included participant demographics, familiarity with GAI tools, frequency and context of usage, perceived effectiveness in teaching and learning, boundaries of AI support, and satisfaction with specific functionalities.

Of the respondents, 66.7\% were undergraduate students, 8.3\% postgraduate, and 8.3\% doctoral candidates, while the remaining participants represented academic staff across various ranks (Lecturer, Assistant Professor, Associate Professor, and Full Professor) as well as independent researchers (Figure~\ref{participants_bacground}).

\begin{wrapfigure}{r}{0.5\textwidth} 
     \centering    
     \hspace{-10mm}
         \begin{tikzpicture}
            \begin{axis}[
    ybar=.05cm,
    every node near coord/.append style={font=\tiny},
    legend style={font=\tiny},
    tick label style={font=\tiny},
    ylabel near ticks, ylabel shift={-6pt},
    width=\linewidth,
    width=6.5cm,
    height=5.5cm,
    every node near coord/.append style={
                        anchor=west,
                        rotate=75
                },
    enlargelimits=.1,
    enlarge y limits={0.1,upper},
    legend style={at={(0.5,-0.32)},
    anchor=north, legend columns=-1},
    ymin=0, 
    ylabel={(\%)},
    xtick={1,2,3,4,5,6,7,8,9,10},
    nodes near coords,
    ytick={0,20,40,60,    80,    100},
    xticklabels={ChatGPT, DeepSeek, Claude, Gemini, Copilot, Other},
    x tick label style={rotate=25,anchor=east},
    grid=both,
    nodes near coords,
    nodes near coords align={vertical},
    bar width=15pt,
    bar shift=0,
    label style={font=\footnotesize},
    ]

\addplot [draw=black, color=orange, preaction={fill, orange}, semithick, pattern color = white]coordinates {(1,100)}; 

\addplot [draw=black, color=Blue, preaction={fill, Blue}, semithick, pattern color = white] coordinates {(2,58.3)}; 

\addplot [draw=black, color=Cyan, preaction={fill, Cyan}, semithick, pattern color = white] coordinates {(3,20.8)}; 

\addplot [draw=black, color=ForestGreen, preaction={fill, ForestGreen}, semithick, pattern color = white] coordinates {(4,54.2)}; 
\addplot [draw=black, color=OrangeRed, preaction={fill, OrangeRed}, semithick, pattern color = white] coordinates {(5,37.5)}; 

\addplot [draw=black, color=teal, preaction={fill, teal}, semithick, pattern color = white] coordinates {(6,16.7)}; 




\end{axis}
 
        \end{tikzpicture}   
        \caption{Familiarity and usage frequency of GAI tools}
        \label{familiarity}       
\end{wrapfigure}
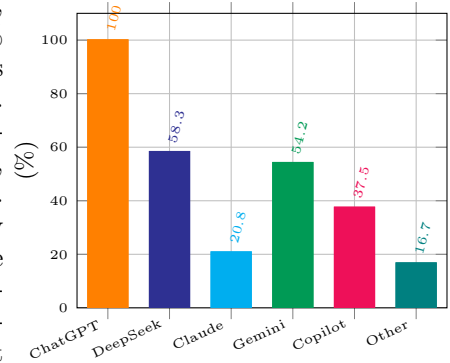

In terms of disciplinary distribution, 79.2\% of respondents were from Computer Science and Engineering, 8.3\% from AI, and another 8.3\% from Science-related fields such as Mathematics, Physics, and Statistics. The remaining participants represented diverse areas including Health, Business, Medicine, and Agriculture. All participants reported familiarity with ChatGPT, while 58.3\% were also acquainted with DeepSeek (Figure~\ref{familiarity}). Moreover, over 95\% of respondents had used GAI tools for at least six months, with usage spanning up to two years. Among these, 87.5\% reported daily interaction with ChatGPT, compared to 20.8\% for DeepSeek.


The survey was structured in two parts. The first part included multiple-choice questions designed to capture participants’ current use of GAI technologies and their corresponding ratings. The second part comprised open-ended questions aimed at collecting qualitative insights into the strengths, challenges, and future prospects of GAI.

\begin{table}[h]
\centering
\caption{Comparison of ChatGPT and DeepSeek based on the highest ratings and largest participant share for each aspect.}\label{portion }
\begin{tabular}{c|c|c|c|c}
\hline \hline
\multirow{2}{*}{\textbf{Aspect}} & \multicolumn{2}{c|}{\textbf{ChatGPT}} & \multicolumn{2}{c}{\textbf{DeepSeek}} \\
\cline{2-5}
 & Best Rating & Max Participants(\%) & Best Rating & Max Participants(\%) \\
\hline \hline
Response Accuracy & 4 & 45.8 & 3 & 45.8 \\
\hline
Relevance to Education & 4 & 58.3 & 4 & 37.5 \\
\hline
Helpful in Research & 4 & 41.7 & 3 & 33.30 \\
\hline
Lecture Preparation & 4 & 36.40 & 3 & 45.50 \\
\hline
Creative Response & 4 & 41.70 & 3 & 45.8 \\
\hline
Coding & 4 & 41.70 & 4 & 50.00 \\

\hline \hline
\end{tabular}

\end{table}

\begin{figure}[h]
     \centering
     \begin{subfigure}[b]{0.48\linewidth}
     \centering
         \captionsetup{justification=centering}
         \begin{tikzpicture}[scale=.90]
            \begin{axis}[
    xlabel={Rating},
    ylabel={Participants (\%)},
    ylabel near ticks, ylabel shift={-5pt},
    width=7cm,
    height=4.5cm,
    grid,
    grid style={gray!50},
    grid=both,
    ymax=65,
    ytick={0,10,20,30,40,50,60,70},
    symbolic x coords={1,2,3, 4, 5},
    xtick=data,
    ymajorgrids=true,
    grid style=solid,
    label style={font=\tiny},
    every node near coord/.append style={font=\tiny},
    legend style={font=\tiny},
    tick label style={font=\tiny},
    legend style={at={(0.5,-0.21)},
    legend style={draw=none},
    anchor=north, legend columns=2},
]


\addplot[mark=otimes*, orange, solid, every mark/.append style={solid, fill=orange}]
    coordinates {
(1,0)
(2,0)
(3,12.5)
(4,45.8)
(5, 41.7)
    };
    \addlegendentry{Response Accuracy}
    
\addplot[mark=square*, blue, solid, every mark/.append style={solid, fill=blue}]
    coordinates {
(1,0)
(2,4.2)
(3,8.3)
(4,58.30)
(5, 29.2)

    };
\addlegendentry{Educational Tasks}

\addplot[mark=diamond*, red, solid, every mark/.append style={solid, fill=red}]
    coordinates {
(1,0)
(2,4.2)
(3,25)
(4,41.7)
(5, 29.2)
    };
\addlegendentry{Helpfulness in Research}

\addplot[mark=halfcircle*, teal, solid, every mark/.append style={solid, fill=teal}]
    coordinates {
(1,0)
(2,13.6)
(3,22.7)
(4,36.4)
(5, 27.3)

    };
\addlegendentry{Lecture Preparation}
    
\addplot[mark=triangle*, magenta, solid, every mark/.append style={solid, fill=magenta}]
    coordinates {
(1,0)
(2,12.5)
(3,20.8)
(4,41.7)
(5, 25)

    };
\addlegendentry{Creativity in Responses}

\addplot[mark=pentagon*, Green, solid, every mark/.append style={solid, fill=Green}]
    coordinates {
(1,0)
(2,0)
(3,25)
(4,41.7)
(5, 33.3)

    };
\addlegendentry{Coding Task}

\end{axis}
        \end{tikzpicture}     
        \caption{ChatGPT}
         \label{training_parameters}
     \end{subfigure}
     \hfill
       \begin{subfigure}[b]{0.48\linewidth}
       \centering
        \captionsetup{justification=centering}
         \begin{tikzpicture}[scale=.90]
            \begin{axis}[
    xlabel={Rating},
    ylabel={Participants (\%)},
    ylabel near ticks, ylabel shift={-5pt},
    width=7cm,
    height=4.5cm,
    grid,
    grid style={gray!50},
    grid=both,
    ymax=65,
    ytick={0,10,20,30,40,50,60,70},
    symbolic x coords={1,2,3, 4, 5},
    xtick=data,
    ymajorgrids=true,
    grid style=solid,
    label style={font=\tiny},
    every node near coord/.append style={font=\tiny},
    legend style={font=\tiny},
    tick label style={font=\tiny},
    legend style={at={(0.5,-0.21)},
    legend style={draw=none},
    anchor=north, legend columns=2},
]


\addplot[mark=otimes*, orange, dashdotted, every mark/.append style={solid, fill=orange}]
    coordinates {
(1,0)
(2,4.2)
(3,45.8)
(4,25)
(5, 25)
    };
    \addlegendentry{Response Accuracy}
    
\addplot[mark=square*, blue, dashdotted, every mark/.append style={solid, fill=blue}]
    coordinates {
(1,0)
(2,8.3)
(3,29.2)
(4,37.5)
(5, 25)

    };
\addlegendentry{Educational Tasks}

\addplot[mark=diamond*, red, dashdotted, every mark/.append style={solid, fill=red}]
    coordinates {
(1,4.2)
(2,8.3)
(3,33.3)
(4,29.2)
(5, 25)
    };
\addlegendentry{Helpfulness in Research}

\addplot[mark=halfcircle*, teal, dashdotted, every mark/.append style={solid, fill=teal}]
    coordinates {
(1,4.5)
(2,9.1)
(3,45.5)
(4,18.2)
(5, 22.7)

    };
\addlegendentry{Lecture Preparation}
    
\addplot[mark=triangle*, magenta, dashdotted, every mark/.append style={solid, fill=magenta}]
    coordinates {
(1,0)
(2,4.2)
(3,45.8)
(4,33.3)
(5, 16.7)

    };
\addlegendentry{Creativity in Responses}

\addplot[mark=pentagon*, Green, dashdotted, every mark/.append style={solid, fill=Green}]
    coordinates {
(1,0)
(2,4.2)
(3,20.8)
(4,50)
(5, 25)

    };
\addlegendentry{Coding Task}

\end{axis}
        \end{tikzpicture}
        
        \caption{DeepSeek}
         \label{time_curve}
     \end{subfigure}
          \caption{Comparison of participant ratings across multiple aspects of ChatGPT and DeepSeek.}

        \label{rating_participants_comparison} 
  
\end{figure}
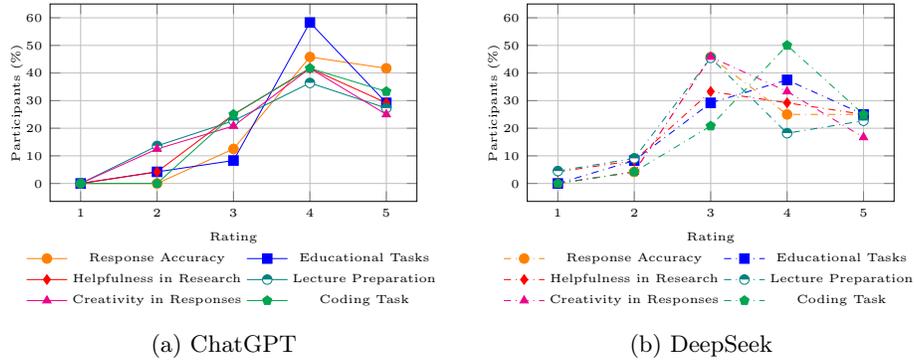

Table \ref{portion } summarizes user evaluations of ChatGPT and DeepSeek across six key aspects relevant to education and research. Participants rated each aspect on a 1–5 scale. For each aspect, we highlight the most frequently selected rating and the percentage of participants who endorsed it. ChatGPT received its strongest rating in \emph{Response Accuracy}, with 45.8\% of respondents assigning a rate of 4, while DeepSeek’s highest concentration in this category was at rating 3, also with 45.8\%. In terms of \emph{Relevance to Education}, ChatGPT was rated 4 by 58.3\% of participants, compared to 37.5\% for DeepSeek. Similarly, in the \emph{Helpfulness in Research} aspect, 41.7\% of respondents gave ChatGPT a rating of 4, whereas 33.3\% rated DeepSeek at 3. For \emph{Lecture Preparation} and \emph{Creative Response}, ChatGPT again scored higher (rating 4) than DeepSeek (rating 3). Interestingly, the trend reversed in the \emph{Coding} aspect, where 50.0\% of participants rated DeepSeek at 4, surpassing ChatGPT’s 41.7\%. Overall, these findings suggest that ChatGPT is perceived as more accurate, creative, and educationally relevant, while DeepSeek demonstrates relative strengths in coding support and certain lecture preparation tasks. Figure \ref{rating_participants_comparison} provides a visual comparison of ratings and participant distributions for each aspect.

\begin{figure} [h]
  \includegraphics[width=1\linewidth]{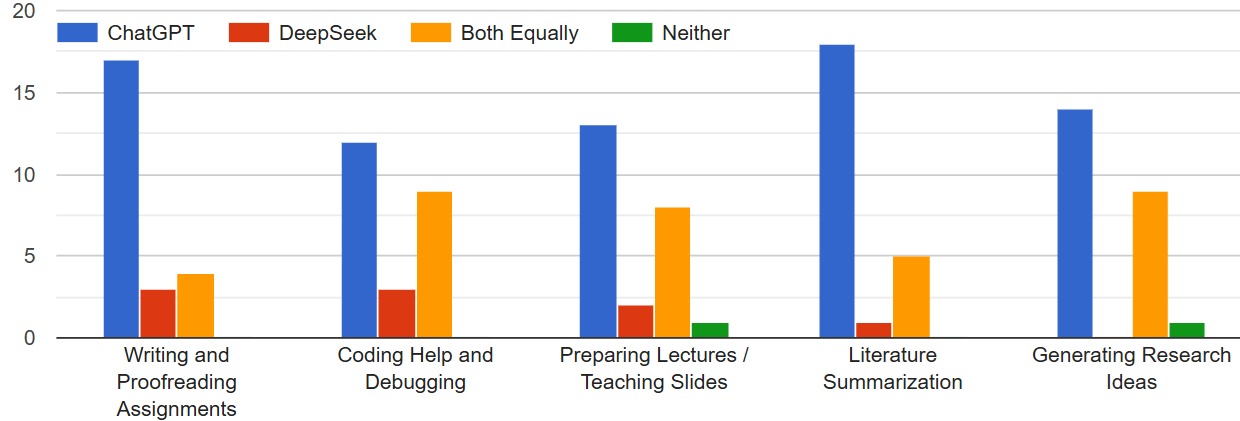}
    \caption{Participants’ perceptions of ChatGPT versus DeepSeek in five academic tasks}
  \label{helpfultool}
\end{figure}

To further assess the practical utility of LLMs, participants were asked: \emph{“Which tool do you find more helpful for the following tasks?”} Figure \ref{helpfultool} summarizes responses across five representative academic activities. For \texttt{Writing and Proofreading Assignments}, the majority of respondents (17) favored ChatGPT, citing its effective phrasing suggestions and grammar correction. By contrast, only 3 preferred DeepSeek, largely for its search-oriented support, while 4 indicated that both tools were equally useful. In \texttt{Coding Help and Debugging}, ChatGPT again led with 12 votes, appreciated for its ability to interpret and explain error messages. DeepSeek received 3 votes for code snippet retrieval, whereas 9 participants considered both tools equally effective. When evaluating \texttt{Preparing Lectures and Teaching Slides}, 13 participants highlighted ChatGPT’s slide-ready summaries as particularly helpful, compared to 2 who preferred DeepSeek. Interestingly, 8 respondents indicated that both tools produced the best results, while 1 found neither helpful. The widest gap was observed in \texttt{Literature Summarization}: 18 participants favored ChatGPT for its concise synthesis of research articles, compared to just 1 for DeepSeek and 5 for both. Finally, in \texttt{Generating Research Ideas}, 14 participants emphasized ChatGPT’s creativity, none selected DeepSeek alone, 9 recognized both tools as equally useful, and 1 found neither beneficial. Overall, these findings indicate ChatGPT’s dominant perceived utility across writing, coding, teaching, summarization, and ideation tasks. 

\begin{wrapfigure}{r}{0.5\textwidth}   
\vspace{-10mm}
  \includegraphics[width=1\linewidth]{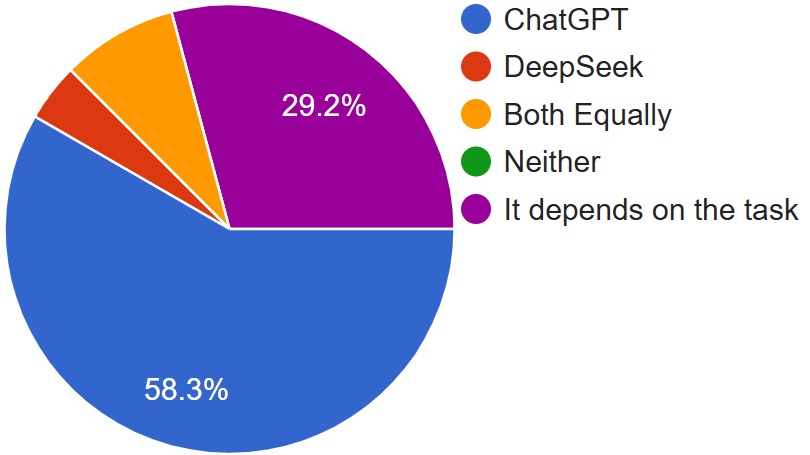}
    \caption{Participants’ perceived reliability of ChatGPT and DeepSeek for education and research.}
  \label{distribution}
\end{wrapfigure}


Figure \ref{distribution} illustrates participants’ perceptions of platform reliability in \texttt{education and research} contexts. A majority of respondents (58.3\%) identified ChatGPT as the most reliable tool. Nearly one-third (29.2\%) emphasized that reliability is task-dependent, suggesting that different platforms may excel in different contexts. A smaller share (8.3\%) rated both tools equally, while  4.2\% considered DeepSeek the most reliable option. Notably, none of the respondents selected \texttt{Neither}. These findings highlight ChatGPT’s strong reputation for reliability, while also underscoring that a meaningful proportion of users value the situational strengths of each platform.

Finally, an open-ended survey revealed key themes regarding the strengths, limitations, and prospects of ChatGPT and DeepSeek. Regarding the \textbf{strengths} of ChatGPT, participants consistently emphasized its versatility, fluency, and speed. Users described it as a \textit{“powerful AI tool, helpful for studies and any query”} and praised its \textit{“early and accurate response of any question.”} Several highlighted its reasoning ability, especially after the introduction of the Deep Research feature, with one noting: \textit{“To me, ChatGPT has been the best when it came to creating a multi-disciplinary research idea.”} It was also valued for generating human-like text, creative content, and programming support. In contrast, responses on DeepSeek were more mixed, reflecting more limited user experience. Those familiar with the tool pointed to its technical accuracy, particularly in coding and data analysis, with one participant remarking that \textit{“the deep thought chain, though it takes more time to answer, is high and precise, especially for code-related issues.”} Others emphasized its accessibility, noting that it \textit{“allows users to insert images or documents without requiring any subscription fee.”} Overall, ChatGPT was viewed as a versatile, user-friendly, and general-purpose assistant, while DeepSeek was recognized for its depth and precision in specialized technical tasks.

When asked about the \textbf{limitations} of ChatGPT and DeepSeek, participants reported a range of challenges related to accuracy, usability, and scope. Several noted that responses can be unreliable, with one participant cautioning that \textit{“sometimes the answer is made up, in particular for searching papers”} and another adding that both models \textit{“failed to include correct references”} when evidence was requested. Logical and numerical errors were also common, as respondents observed that \textit{“both make mistakes in arithmetic operations”} and that ChatGPT is \textit{“very poor for low-level language [and] sometimes makes mistakes in calculations.”} Usability concerns included slower responses during extended conversations and the tendency to forget context when switching windows, while others pointed out that \textit{“it talks too much; sometimes you would want a short response.”} DeepSeek, in particular, was described as \textit{“time consuming”} in generating answers. Broader limitations such as lack of social or emotional intelligence, difficulty handling less popular questions, and restricted access without premium features were also cited. Collectively, these comments highlight that while both tools are powerful, users remain aware of their vulnerabilities in accuracy, efficiency, and contextual understanding. 

We also asked, \textit{“Do you think these tools will transform education and research in the future? If so, how?”}, most participants responded affirmatively, often noting that the impact is already visible. Several emphasized their role in enhancing productivity, with one respondent stating that \textit{“the research productivity has already tremendously increased”} and another adding that \textit{“the future of education must be combined with AI.”} Many highlighted benefits for students, including immediate access to explanations—\textit{“if students get any problem they can immediately ask it in these tools”}—as well as support for generating research topics and ideas. Others underscored the potential for personalized learning and more efficient data analysis. At the same time, a few participants expressed caution, warning that overreliance on AI could diminish critical thinking, as reflected in the concern that \textit{“the more people are getting familiar with them, their own brain cells are becoming. People are to regret ultimately, no matter whatsoever.”} Overall, the consensus suggests that generative AI is widely expected to play a transformative role in education and research, though its long-term impact will depend on addressing current weaknesses and ensuring responsible use.

\section{Conclusion}

 In this study, we conducted a systematic investigation into the technical foundations, application performances, and user perceptions of two state-of-the-art GAI models: ChatGPT and DeepSeek. Our research was structured around three core objectives. First, we performed a detailed comparative analysis of the technical architectures, capabilities, and design principles underlying ChatGPT and DeepSeek, highlighting their respective strengths, unique features, and innovations in GAI development. Second, we executed a series of extensive generative experiments utilizing both models across diverse application domains. The generated outputs were rigorously validated through assessments by domain-specific human experts to ensure accuracy, relevance, and practical applicability of the models responses. Third, we conducted a comprehensive user survey to capture practical experiences, usability feedback, and satisfaction levels of end-users who regularly engage with ChatGPT and DeepSeek across various tasks. In addition to these empirical evaluations, we systematically reviewed and synthesized the growing body of literature related to ChatGPT and DeepSeek, summarizing key findings, emerging application trends, and documented challenges associated with the deployment of GAI systems. Our study contributes to a deeper understanding of how GAI models are being adopted in real-world settings, provides insights into their evolving capabilities, and highlights the ongoing need for responsible development, evaluation, and governance as these technologies continue to mature and impact society at scale.



\bibliographystyle{splncs04}

\end{document}